\documentclass[conference]{IEEEtran}
\IEEEoverridecommandlockouts
\usepackage{cite}
\usepackage{amsmath,amssymb,amsfonts}
\usepackage{algorithmic}
\usepackage{graphicx}
\usepackage{textcomp}
\usepackage{xcolor}
\usepackage{booktabs}
\usepackage{multirow}
\usepackage{dblfloatfix}
\usepackage{subcaption}  
\usepackage{booktabs,subcaption}
\usepackage{upgreek}
\makeatletter
\newcommand{\linebreakand}{%
  \end{@IEEEauthorhalign}
  \hfill\mbox{}\par
  \mbox{}\hfill\begin{@IEEEauthorhalign}
}
\makeatother

\def\BibTeX{{\rm B\kern-.05em{\sc i\kern-.025em b}\kern-.08em
    T\kern-.1667em\lower.7ex\hbox{E}\kern-.125emX}}
\begin{document}

\title{Unsupervised Domain Adaptation with SAM-RefiSeR for Enhanced Brain Tumor Segmentation
}

\author{
\IEEEauthorblockN{Dillan Imans$^\ast$}
\thanks{$^\ast$Equal contributions, $^\dag$Corresponding author: choo@skku.edu}
\IEEEauthorblockA{\textit{Dept. of Computer Science and Engineering} \\
\textit{Sungkyunkwan University, South Korea}}
\and
\IEEEauthorblockN{Phuoc-Nguyen Bui$^\ast$}
\IEEEauthorblockA{\textit{Research Convergence Institute} \\
\textit{Sungkyunkwan University, South Korea}}
\linebreakand
\IEEEauthorblockN{Duc-Tai Le}
\IEEEauthorblockA{\textit{Dept. of Electrical and Computer Engineering} \\
\textit{Sungkyunkwan University, South Korea}}
\and
\IEEEauthorblockN{Hyunseung Choo$^\dag$}
\IEEEauthorblockA{\textit{Dept. of Electrical and Computer Engineering} \\
\textit{Sungkyunkwan University, South Korea}}
}
\maketitle

\begin{abstract}
Robust brain-tumor segmentation in MRI must withstand domain shifts from heterogeneous scanners and protocols. Unsupervised domain adaptation (UDA) can exploit plentiful unlabeled data, yet many approaches erode boundaries, suppress tumor cues, or amplify errors from noisy pseudo-labels. General-purpose models like SAM also struggle on MRI due to the domain gap and a lack of morphology-aware consistency.
We propose \textbf{SAM-RefiSeR}, a two-phase UDA framework that integrates SAM for reliable, annotation-efficient segmentation. Phase I reduces source–target discrepancy via Fourier-based frequency adaptation and adversarial feature alignment, aligning appearance and representations while preserving anatomy. Phase II adopts a student–teacher scheme in which SAM first refines pseudo-labels, then gates them with confidence- and morphology-aware criteria to suppress unreliable masks and curb error propagation.
Across diverse cross-modality settings, SAM-RefiSeR consistently surpasses strong UDA baselines, particularly under severe shifts. By improving boundary fidelity and robustness without additional labels, SAM-RefiSeR brings brain-tumor segmentation closer to practical, generalizable clinical deployment.
\end{abstract}


\section{Introduction}

Deep neural networks excel in brain-tumor segmentation when train and test conditions match, but accuracy collapses under domain shift. Small variations in scanner type, field strength, or protocol can significantly alter image statistics, undermining single-domain models \cite{bui2023semi}. Annotating masks for every acquisition protocol is impractical, while unlabeled MRIs are plentiful and often shareable once de-identified.
Unsupervised domain adaptation (UDA) addresses this by leveraging a labeled source domain and an unlabeled target domain to achieve domain-robust representations without additional. UDA methods fall into three categories: Pixel-level adaptation (e.g., CycleGAN \cite{zhu2017unpaired}) aligns appearance statistics but may distort fine tumor boundaries; Feature-level alignment (e.g., ADVENT \cite{vu2019advent}) enforces domain-invariant embeddings, but can suppress tumor-specific cues; Output-level adaptation (e.g., CBST \cite{zou2018unsupervised}) leverages pseudo-labels but suffers from cumulative label noise. Despite enhancing cross-domain consistency, these trade-offs still limit performance.

Foundation models such as SAM~\cite{kirillov2023segment} offer broad segmentation capacity from large-scale natural-image pretraining but struggle with the MRI domain gap, falling short of specialized medical architectures. Adapters (e.g., \cite{chen2023sam}) improve transfer but often lack morphology-aware volumetric consistency essential for reliable tumor boundaries. We propose \textbf{SAM-RefiSeR}, a two-phase UDA framework: Phase~I combines adversarial and Fourier-based alignment to reduce source-target discrepancy while preserving anatomy; Phase~II employs a student-teacher loop where the teacher generates prototype pseudo-labels and SAM refines/gates them via confidence and morphology tests, curbing error propagation and enabling the student to surpass SAM on delineation. The main contributions are:

\begin{itemize}
\item We present SAM-RefiSeR, a two-phase UDA pipeline for MRI brain–tumor segmentation that couples adversarial and frequency alignment with a SAM-guided student–teacher refinement to sharpen pseudo-labels.
\item We propose a novel pseudo-label curation strategy that integrates SAM-derived confidence metrics and morphology-based selection to retain anatomically plausible masks, mitigating cumulative errors and enhancing target-domain robustness.
\item Extensive experiments show SAM-RefiSeR consistently outperforming strong UDA baselines.
\end{itemize}

\begin{figure*}[!t]
    \centering
    \includegraphics[width=\textwidth]{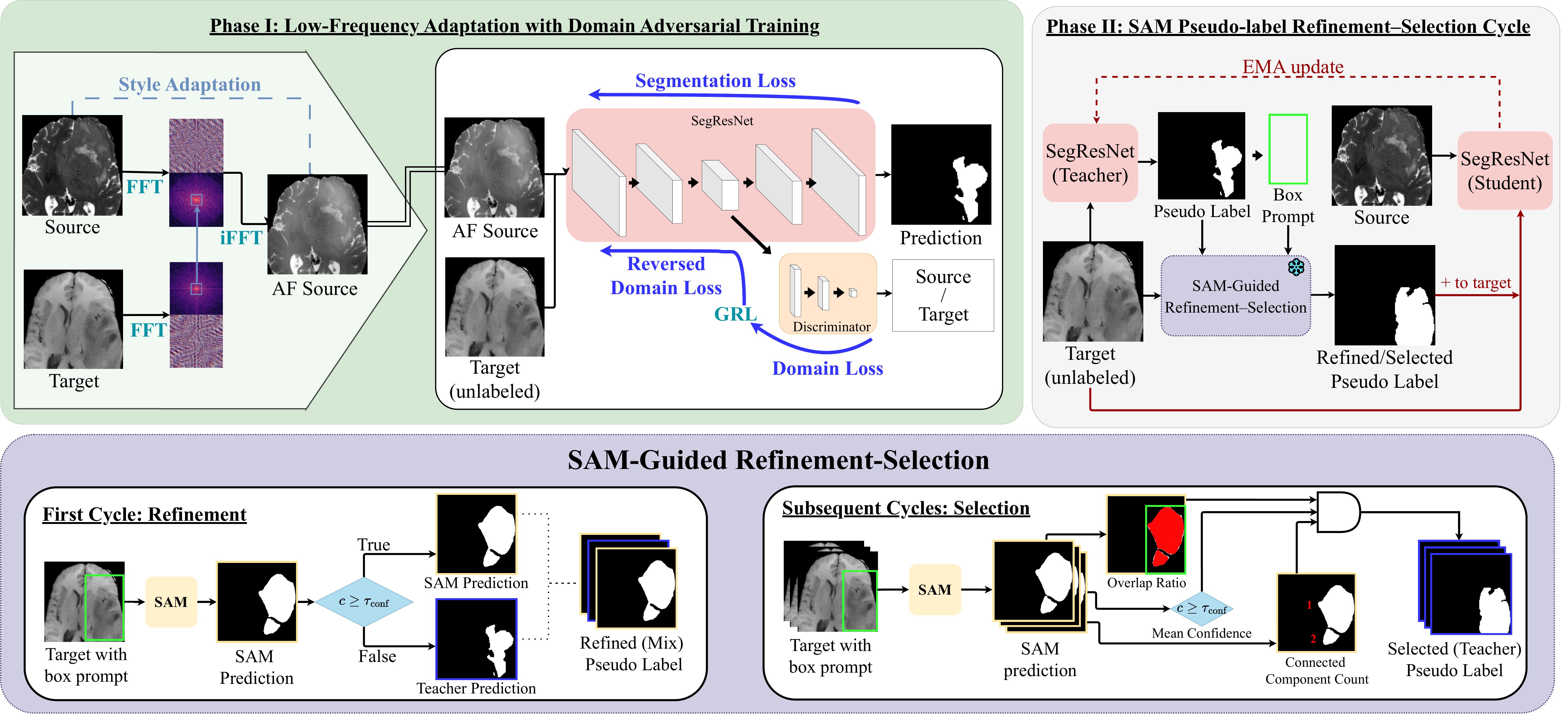}
\caption{Overview of the proposed 2-phase framework SAM-RefiSeR. (i)FFT, GRL, EMA, AF stand for (inverse) Fast Fourier Transform, Gradient Reversal Layer, Exponential Moving Average, and Amplitude Fused, respectively.}

    \label{fig:main-figure}
\end{figure*}


\section{Methods}

The overall architecture of \textsc{SAM-RefiSeR} is illustrated in Fig.~\ref{fig:main-figure} and unfolds in two consecutive phases.


\subsection{Phase I: Low-Frequency Adaptation with Domain Adversarial Training}
\textbf{Frequency adaptation \cite{yang2020fda}} Each source volume \(x^{s}\) is paired with a randomly selected target volume \(x^{t}\). Both are mapped to the spectral domain via a 3D fast Fourier transform (FFT), \(\hat{x}^{s}=\operatorname{FFT}(x^{s})\) and \(\hat{x}^{t}=\operatorname{FFT}(x^{t})\). All spectra are FFT‑shifted so that the zero‑frequency component lies at the array center.
The source spectrum is factorized into amplitude and phase, \(A^{s}=|\hat{x}^{s}|\) and \(\Phi^{s}=\angle\hat{x}^{s}\), while only the amplitude \(A^{t}=|\hat{x}^{t}|\) is retained from the target.  
Let \(L\in(0,1)\) be a scalar that sets the cube size and define 
\(b=\lfloor L\,\min(D,H,W)\rfloor\). Empirically, $L\!\in\![0.01,0.03]$ works best across MRI.
With the volume center at \((D/2,\,H/2,\,W/2)\), the low-frequency cube is as follows:
\[
\mathcal{B}=\bigl\{(d,h,w)\,\bigl|\,
      |d-\tfrac{D}{2}|\le b,\;
      |h-\tfrac{H}{2}|\le b,\;
      |w-\tfrac{W}{2}|\le b\bigr\}.
\]
For all $(d,h,w)\!\in\!\mathcal{B}$, set $A^s_{d,h,w}\!\leftarrow\!A^t_{d,h,w}$. Swapping only low-frequency coefficients (style: intensity/contrast) while keeping high-frequency detail transfers appearance without blurring~\cite{yang2020fda}.
Finally, we fuse the modified amplitude \(A^{s\leftarrow t}\) with the original phase \(\Phi^{s}\) and apply an inverse FFT,
\(
\tilde{x}^{s} = \operatorname{iFFT}\bigl(A^{s\leftarrow t} e^{i\Phi^{s}}\bigr), 
\)
where \(A^{s\leftarrow t}\) takes target amplitudes inside \(\mathcal{B}\) and source amplitudes elsewhere. The resulting volume \(\tilde{x}^{s}\) retains source anatomy while adopting target-style appearance.

\begin{table*}[!b]
  \centering
\caption{Comparison of methods across imaging modalities: T1CE as source (left) and T2 as source (right). \\ ↑: higher is better; ↓: lower is better.}
\label{tab:main_benchmark_results}
  \setlength{\tabcolsep}{4pt}
  \renewcommand{\arraystretch}{0.93}
  \begin{subtable}[t]{0.48\textwidth}
    \centering
    \caption{T1CE as source}
    \begin{tabular}{lcccccc}
      \toprule
      \multirow{2}{*}{\textbf{Method}} & \multicolumn{2}{c}{\textbf{T2}} & \multicolumn{2}{c}{\textbf{FLAIR}} & \multicolumn{2}{c}{\textbf{T1}} \\
      \cmidrule(lr){2-3}\cmidrule(lr){4-5}\cmidrule(lr){6-7}
      & \textbf{Dice\,↑} & \textbf{HD95\,↓} & \textbf{Dice\,↑} & \textbf{HD95\,↓} & \textbf{Dice\,↑} & \textbf{HD95\,↓} \\
      \midrule
      Baseline
        & 7.6  & 76.4  & 28.9 & 71.3 & 19.9 & 59.7  \\
      \midrule
      DAR‑UNET\cite{yao2022darunet}
        & 35.5 & 57.1  & 43.2 & 40.3 & 26.3 & 51.1 \\
      \midrule
        DAFormer\cite{hoyer2022daformer}
          & 41.0 & 43.3 & 50.4 & 27.2 & 49.8 & 25.3 \\
        \midrule
        MIC\cite{hoyer2023mic}
          & 45.2 & 39.2 & 49.3 & 28.3 & 49.4 & 24.3 \\
        \midrule
        HRDA\cite{hoyer2022hrda}
          & 34.4 & 49.5  & 51.1 & 26.1 & 49.4 & 26.0 \\
        \midrule
        
     Ours
        & \textbf{54.2} & \textbf{25.3} & \textbf{58.5} & \textbf{19.6} & \textbf{57.1} & \textbf{23.2} \\
      \bottomrule
    \end{tabular}%
  \end{subtable}
   \hfill
  \begin{subtable}[t]{0.48\textwidth}
    \centering
    \caption{T2 as source}
    \begin{tabular}{lcccccc}
      \toprule
      \multirow{2}{*}{\textbf{Method}} & \multicolumn{2}{c}{\textbf{T1CE}} & \multicolumn{2}{c}{\textbf{FLAIR}} & \multicolumn{2}{c}{\textbf{T1}} \\
      \cmidrule(lr){2-3}\cmidrule(lr){4-5}\cmidrule(lr){6-7}
      & \textbf{Dice\,↑} & \textbf{HD95\,↓} & \textbf{Dice\,↑} & \textbf{HD95\,↓} & \textbf{Dice\,↑} & \textbf{HD95\,↓} \\
      \midrule
      Baseline
        & 8.5  & 64.5  & 55.6 & 23.6 & 4.7  & 80.2  \\
      \midrule
      DAR‑UNET\cite{yao2022darunet}
        & 34.9 & 51.8  & 45.4 & 31.2 & 15.8 & 76.3  \\
      \midrule
    DAFormer\cite{hoyer2022daformer}
      & 48.0 & 40.3 & 59.8 & 20.6 & 45.6 & 34.3 \\
    \midrule
    MIC\cite{hoyer2023mic}
      & 52.6 & 36.0 & 60.8 & 20.2 & 40.7 & 37.3 \\
    \midrule
    HRDA\cite{hoyer2022hrda}
      & 52.9 & 27.7 & 62.8 & 19.3 & 43.5 & 29.3 \\
    \midrule
    Ours
        & \textbf{60.9} & \textbf{23.5} & \textbf{64.4} & \textbf{16.3} & \textbf{50.2} & \textbf{24.6} \\
      \bottomrule
    \end{tabular}%
  \end{subtable}
\end{table*}

\begin{figure*}[!b]
  \centering
  \begin{subfigure}[t]{0.49\textwidth}
    \centering
    \includegraphics[width=\linewidth]{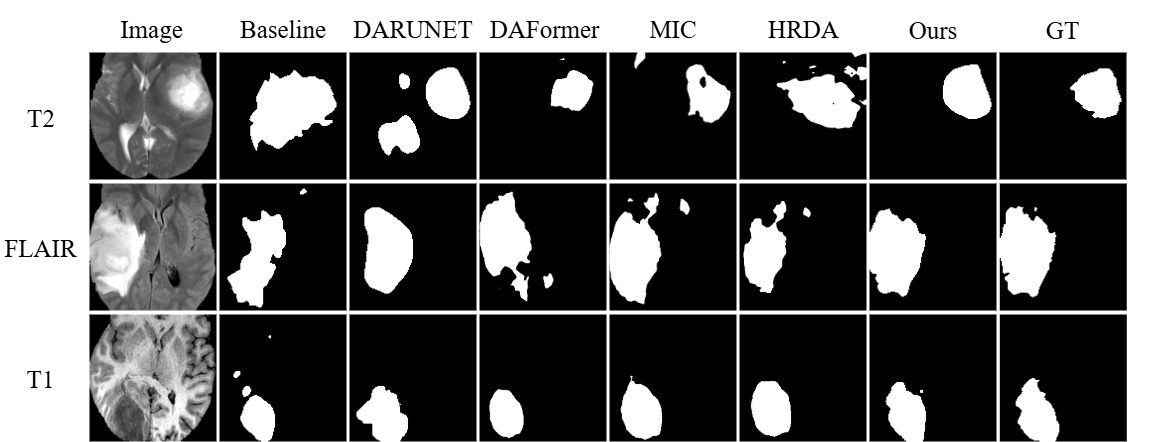}
    \caption{Model trained on T1CE.}
    \label{fig:qual_t1ce}
  \end{subfigure}%
  \hfill
  \begin{subfigure}[t]{0.49\textwidth}
    \centering
    \includegraphics[width=\linewidth]{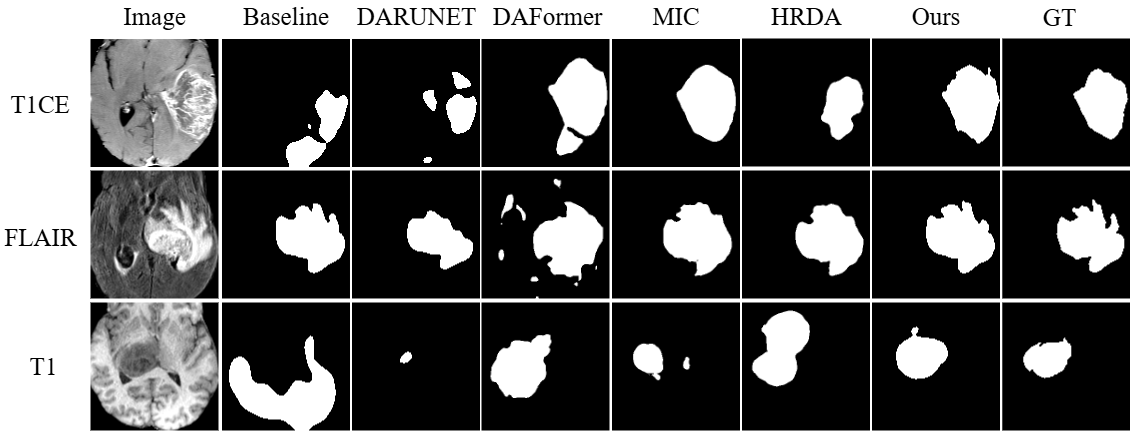}
    \caption{Model trained on T2.}
    \label{fig:qual_t2}
  \end{subfigure}
  \caption{Qualitative comparison of methods with T1CE and T2 as source domains.}
  \label{fig:qual_column}
\end{figure*}

\textbf{Domain adversarial training} We promote domain-invariant features using DANN~\cite{ganin2016domain}. Let $f_{\theta}$ be the segmentation encoder and $g_{\psi}$ the domain discriminator predicting $d\!\in\!\{0,1\}$ (source/target) from $z=f_{\theta}(x)$. The encoder and discriminator are trained adversarially via backprop to minimize the loss as follows:

\begin{align}
\min_{\theta}\;&\nonumber
    \mathcal{L}_{\text{seg}}\!\bigl(f_{\theta}(x^{s}), y^{s}\bigr)
  + \lambda\,\mathcal{L}_{\text{dom}}\!\bigl(g_{\psi}(f_{\theta}(x)), d\bigr),\\
\min_{\psi}\;&\nonumber
    \mathcal{L}_{\text{dom}}\!\bigl(g_{\psi}(f_{\theta}(x)), d\bigr).
\end{align}

We use soft Dice~\cite{milletari2016v} on labeled source images for $\mathcal{L}_{\text{seg}}$ and binary cross-entropy over both domains for $\mathcal{L}_{\text{dom}}$. Adversarial learning employs a GRL~\cite{ganin2016domain} between encoder $f_{\theta}$ and discriminator $g_{\psi}$. To avoid early instability we use a two-phase schedule: a short warm-up with $\lambda\!=\!0$, then a logistic ramp to $\lambda_{\max}$, after which $\lambda$ is fixed.
\[
\lambda(t)=\frac{\lambda_{\max}}{1+\exp\!\bigl(-\gamma\,(t-t_{0})\bigr)},
\]
$t$ is the epoch, $t_{0}$ the ramp midpoint, $\gamma$ controls its steepness.

\subsection{Phase II: SAM‑Guided Pseudo‑label Refinement--Selection}


Using the model trained in Phase~I as the teacher, we initiate iterative self-training on the unlabeled target domain.

\textbf{First cycle: SAM‑guided pseudo‑label refinement.}
Let the warm‑up teacher produce an initial 3D pseudo‑label volume
\(y_i^{(0)} \in \{0,1\}^{D \times H \times W}\) for target case~$i$ (henceforth the superscript $(t)$ denotes the adaptation cycle:
$t=0$ for the warm‑up teacher output, $t=1$ after the present refinement step,
and $t=2,\dots,N$ for the ensuing selection cycles).
Writing \(y_{ij}^{(0)}\in\{0,1\}^{H\times W}\) for its \(j^{\mathrm{th}}\) axial slice,
we compute the minimal enclosing rectangle of its foreground pixels and denote
the resulting binary mask as
\(
B_{ij}\;\subset\;\mathbb{Z}^{H\times W},
\)
which serves as the bounding‑box prompt for SAM. We then invoke SAM on that mask \(B_{ij}\) to obtain a refined slice‑mask \(s_{ij}^{(1)}\) with confidence score \(c_{ij}^{(1)}\). Define the slice‑wise refined label as:
\begin{align}\nonumber
\tilde y_{ij}^{(1)} \;=\;
\begin{cases}
s_{ij}^{(1)}, & c_{ij}^{(1)} \ge \tau_{\mathrm{conf}},\\[4pt]
y_{ij}^{(0)}, & c_{ij}^{(1)} < \tau_{\mathrm{conf}},
\end{cases}
\end{align}

where \(\tau_{\mathrm{conf}}\in(0,1)\) is a global confidence threshold.
Thus each slice is labeled either by SAM (high-confidence case) or by the original
teacher mask. Stacking the slice-level labels yields the refined 3-D volume
\(\tilde y_i^{(1)}=\{\tilde y_{ij}^{(1)}\}_{j=1}^{D}\).

Collecting all target cases gives
\(\tilde{Y}^{(1)}=\{\tilde y_i^{(1)}\}_{i=1}^{n_t}\). We train a student model on \((X_s,Y_s)\cup(X_t,\tilde{Y}^{(1)})\) using standard segmentation loss.  After each epoch, the student weights \(\theta_S\) update the teacher weights  \(\theta_T\) via EMA:
\[
  \theta_T \leftarrow \alpha\,\theta_T + (1-\alpha)\,\theta_S,
  \quad \alpha\in(0,1).
\]

\textbf{Subsequent cycles: SAM‑guided pseudo‑label selection.}
For each cycle \(t\in\{2,\dots,N\}\) the updated teacher predicts
\(y_i^{(t)}\in\{0,1\}^{D\times H\times W}\).
We use SAM to obtain masks \(s_{ij}^{(t)}\) and confidences \(c_{ij}^{(t)}\). The SAM masks are then evaluated with three volume‑level metrics; the teacher pseudo‑labels are left unchanged. Let \(J_i=\{j\mid B_{ij}\text{ exists}\}\) be the slice indices, and
write \(\lvert\cdot\rvert\) for the number of foreground voxels in a binary mask.

\begin{enumerate}
\item Mean confidence
      $\bar c_i^{(t)} = \frac{1}{|J_i|}\sum_{j\in J_i} c_{ij}^{(t)}$. High \(\bar c_i^{(t)}\) signals strong epistemic certainty in SAM’s slice‑wise
      proposals, whereas low values often coincide with weak edges or out‑of‑plane
      artifacts.

\item Overlap ratio
      $o_i^{(t)} =
      \frac{\sum_{j\in J_i} \lvert s_{ij}^{(t)}\rvert}
           {\sum_{j\in J_i} \lvert B_{ij}\rvert},$
      where $\lvert B_{ij}\rvert$ is the number of pixels in the bounding box. Very small ratios flag masks dominated by false positives; ratios near $1$ indicate overly loose boxes capturing non-tumor tissue; intermediate values typically correspond to well-focused, accurate segmentation.

\item Connected‑component  $n_i^{(t)} =
      \operatorname{CC}\Bigl(
        \mathrm{Stack}\bigl\{s_{ij}^{(t)}\bigr\}_{j\in J_i}
      \Bigr),$
      counts connected foreground components in 3D masks formed by stacking all predictions. Masks with a small handful of components are plausible, whereas many tiny components indicate noise or over-segmentation.
\end{enumerate}

A volume is retained iff:
\[
\bar c_i^{(t)} \ge \tau_{\mathrm{conf}}
\;\land\;
o_i^{(t)} \in [\tau_{\mathrm{overlap}}^{-},\,\tau_{\mathrm{overlap}}^{+}]
\;\land\;
n_i^{(t)} \le \tau_{\mathrm{cc}}.
\]

\section{Experiments and Results}

\subsection{Dataset and Implementation Details}
We use the BraTS~2021 multimodal MRI dataset~\cite{menze2014multimodal} with voxel labels for ET (Enhancing Tumor), TC (Tumor Core), and WT (Whole Tumor). We perform a patient-wise 80/20 train/test split. We crop all images to the nonzero region, to $128\times128\times128$, and apply per-volume min--max normalization.
Experiments use PyTorch on two NVIDIA RTX~2080\,Ti. Phase~I trains for 150 epochs with Adam (lr $=10^{-4}$). Phase~II comprises five student-teacher self-training cycles. The teacher is updated by EMA with decay $\alpha=0.99$. Refinement/selection thresholds are chosen via grid search on a small unlabeled target subset: $\tau_{\mathrm{conf}}\!\in\!\{0.5,0.6,\dots,0.9\}$ for both phases; for selection we additionally sweep $\tau_{\mathrm{overlap}}^-\!\in\!\{0.3,0.4,0.5\}$, $\tau_{\mathrm{overlap}}^+\!\in\!\{0.6,0.7,0.8\}$ with $\tau_{\mathrm{overlap}}^-<\tau_{\mathrm{overlap}}^+$, and component caps $\tau_{\mathrm{cc}}\!\in\!\{1,3,5,10,20,30,50\}$.

\subsection{Comparison with Existing Methods}

Tables~\ref{tab:main_benchmark_results} (a,b) report results using T1CE and T2 as source domains, comparing SAM-RefiSeR with SegResNet~\cite{myronenko20183d} and four UDA baselines. Averaged over ET/TC/WT, SAM-RefiSeR achieves the highest Dice and lowest HD95 in both settings.
Notably, SAM-RefiSeR shows the largest improvements under severe domain shifts, such as in T1CE$\rightarrow$T2 (Dice $54.2\%$, HD95 $25.3$ voxels). Improvements persist but narrow in visually closer domains (e.g., T2$\rightarrow$FLAIR). Figures~\ref{fig:qual_column} (a,b) visually demonstrate fewer false positives aligning with the quantitative results, where SAM-RefiSeR accurately delineates subtle tumor structures missed by other methods.

\subsection{Ablation Studies}
Table~\ref{tab:phase_res} ablates Phase~I and Phase~II. Phase~I provides initial improvements in Dice and HD95, while Phase~II delivers the most significant gains, further enhancing performance.
Table~\ref{tab:cycle_res}, which ablates the number of cycles, shows that most gains arrive by cycle~3. Thus, early cycles deliver the bulk of improvement; later cycles offer modest polish.

\begin{table}[!h]
\centering
\caption{Ablation study of model stages from the T1CE- and T2-source domains to different MRI modalities.}
\label{tab:phase_res}
\setlength{\tabcolsep}{6pt}
\renewcommand{\arraystretch}{0.7}
\begin{tabular}{lcccccc}
\toprule
\multicolumn{7}{c}{\textbf{(a) T1CE as source}} \\
\midrule
\multirow{2}{*}{\textbf{Method}} 
  & \multicolumn{2}{c}{\textbf{T2}} 
  & \multicolumn{2}{c}{\textbf{FLAIR}} 
  & \multicolumn{2}{c}{\textbf{T1}} \\
\cmidrule(lr){2-3}\cmidrule(lr){4-5}\cmidrule(lr){6-7}
& Dice$\uparrow$ & HD95$\downarrow$ & Dice$\uparrow$ & HD95$\downarrow$ & Dice$\uparrow$ & HD95$\downarrow$ \\
\midrule
Baseline
  & 7.6  & 76.4  & 28.9 & 71.3 & 19.9& 59.7  \\
\midrule
Phase I
  & 14.6 & 62.7  & 37.5 & 42.0 & 30.1 & 42.8 \\
\midrule
Ours
  & \textbf{54.2} & \textbf{25.3} & \textbf{58.5} & \textbf{19.6} & \textbf{57.1} & \textbf{23.2} \\
\addlinespace[0.5ex]
\specialrule{\heavyrulewidth}{0pt}{0pt}
\addlinespace[0.5ex]

\multicolumn{7}{c}{\textbf{(b) T2 as source}} \\
\midrule
\multirow{2}{*}{\textbf{Method}} 
  & \multicolumn{2}{c}{\textbf{T1CE}} 
  & \multicolumn{2}{c}{\textbf{FLAIR}} 
  & \multicolumn{2}{c}{\textbf{T1}} \\
\cmidrule(lr){2-3}\cmidrule(lr){4-5}\cmidrule(lr){6-7}
& Dice$\uparrow$ & HD95$\downarrow$ & Dice$\uparrow$ & HD95$\downarrow$ & Dice$\uparrow$ & HD95$\downarrow$ \\
\midrule
Baseline
  & 8.5  & 64.5  & 55.6 & 23.6 & 4.7 & 80.2  \\
\midrule
Phase I
  & 27.1 & 43.5  & 58.1 & 19.5 & 19.5 & 73.7  \\
\midrule
Ours
  & \textbf{60.9} & \textbf{23.5} & \textbf{64.4} & \textbf{16.3} & \textbf{50.2} & \textbf{24.6} \\
\bottomrule
\end{tabular}%
\end{table}

\begin{table}[!h]
\centering
\caption{Ablation study of cycle‑wise performance from the T1CE‑ and T2‑source domains to different MRI modalities.}
\label{tab:cycle_res}
\setlength{\tabcolsep}{6pt}
\renewcommand{\arraystretch}{0.7}
\begin{tabular}{cc rr rr rr}
\toprule
\multicolumn{7}{c}{\textbf{(a) T1CE as source}} \\
\midrule
\multirow{2}{*}{\textbf{Cycle}} 
  & \multicolumn{2}{c}{\textbf{T2}} 
  & \multicolumn{2}{c}{\textbf{FLAIR}} 
  & \multicolumn{2}{c}{\textbf{T1}} \\
\cmidrule(lr){2-3}\cmidrule(lr){4-5}\cmidrule(lr){6-7}
& Dice$\uparrow$ & HD95$\downarrow$ & Dice$\uparrow$ & HD95$\downarrow$ & Dice$\uparrow$ & HD95$\downarrow$ \\
\midrule
\textbf{1}
  & 34.7 & 49.7 & 46.1 & 34.2 & 53.1 & 28.5 \\
\midrule
\textbf{3}
  & 47.3 & 30.7 & 55.1 & 20.7 & 56.0 & 26.5 \\
\midrule
\textbf{5}
  & \textbf{54.2} & \textbf{25.3} & \textbf{58.5} & \textbf{19.6} & \textbf{57.1} & \textbf{23.2} \\
\addlinespace[0.5ex]
\specialrule{\heavyrulewidth}{0pt}{0pt}
\addlinespace[0.5ex]
\multicolumn{7}{c}{\textbf{(b) T2 as source}} \\
\midrule
\multirow{2}{*}{\textbf{Cycle}} 
  & \multicolumn{2}{c}{\textbf{T1CE}} 
  & \multicolumn{2}{c}{\textbf{FLAIR}} 
  & \multicolumn{2}{c}{\textbf{T1}} \\
\cmidrule(lr){2-3}\cmidrule(lr){4-5}\cmidrule(lr){6-7}
& Dice$\uparrow$ & HD95$\downarrow$ & Dice$\uparrow$ & HD95$\downarrow$ & Dice$\uparrow$ & HD95$\downarrow$ \\
\midrule
\textbf{1}
  & 51.9 & 39.2 & 62.4 & 19.5 & 33.0 & 39.0 \\
\midrule
\textbf{3}
  & 55.8 & 35.9 & 63.2 & 16.5 & 48.4 & 25.1 \\
\midrule
\textbf{5}
  & \textbf{60.9} & \textbf{23.5} & \textbf{64.4} & \textbf{16.3} & \textbf{50.2} & \textbf{24.6} \\
\bottomrule
\end{tabular}%
\end{table}


\section{Conclusion}
SAM-RefiSeR delivers a robust, two-phase solution to unsupervised domain adaptation for brain-tumor segmentation. The first phase combines Fourier-based frequency adaptation with adversarial training to bridge source–target gaps while preserving anatomical detail. The second phase employs an iterative student–teacher framework, where SAM refines and filters pseudo-labels through confidence and morphology checks, halting error propogation and steadily improving target-domain performance. SAM-RefiSeR consistently outperforms prior UDA baselines across modality shifts, with the largest margins under the most severe cross-modality gaps. Future work will extend evaluations to diverse datasets to validate generalizability and robustness.

\section*{Acknowledgement}
This work was supported in part by IITP grant funded by the Korean government (MSIT) under IITP-2025-RS-2020-II201821 (30\%), RS-2024-00459512 (30\%), RS-2021-II212068 (20\%), and RS-2019-II190421 (20\%).

\bibliographystyle{IEEEtran}
\bibliography{references}

@inproceedings{zhu2017unpaired,
  title={Unpaired image-to-image translation using cycle-consistent adversarial networks},
  author={Zhu, Jun-Yan and Park, Taesung and Isola, Phillip and Efros, Alexei A},
  booktitle={Proceedings of the IEEE international conference on computer vision},
  pages={2223--2232},
  year={2017}
}

@inproceedings{yang2020fda,
  title={Fda: Fourier domain adaptation for semantic segmentation},
  author={Yang, Yanchao and Soatto, Stefano},
  booktitle={Proceedings of the IEEE/CVF conference on computer vision and pattern recognition},
  pages={4085--4095},
  year={2020}
}

@article{ganin2016domain,
  title={Domain-adversarial training of neural networks},
  author={Ganin, Yaroslav and Ustinova, Evgeniya and Ajakan, Hana and Germain, Pascal and Larochelle, Hugo and Laviolette, Fran{\c{c}}ois and March, Mario and Lempitsky, Victor},
  journal={Journal of machine learning research},
  volume={17},
  number={59},
  pages={1--35},
  year={2016}
}

@inproceedings{zou2018unsupervised,
  title={Unsupervised domain adaptation for semantic segmentation via class-balanced self-training},
  author={Zou, Yang and Yu, Zhiding and Kumar, BVK and Wang, Jinsong},
  booktitle={Proceedings of the European conference on computer vision (ECCV)},
  pages={289--305},
  year={2018}
}

@inproceedings{kirillov2023segment,
  title={Segment anything},
  author={Kirillov, Alexander and Mintun, Eric and Ravi, Nikhila and Mao, Hanzi and Rolland, Chloe and Gustafson, Laura and Xiao, Tete and Whitehead, Spencer and Berg, Alexander C and Lo, Wan-Yen and others},
  booktitle={Proceedings of the IEEE/CVF international conference on computer vision},
  pages={4015--4026},
  year={2023}
}

@InProceedings{hoyer2023mic,
  title={{MIC}: Masked Image Consistency for Context-Enhanced Domain Adaptation},
  author={Hoyer, Lukas and Dai, Dengxin and Wang, Haoran and Van Gool, Luc},
  booktitle={Proceedings of the IEEE/CVF Conference on Computer Vision and Pattern Recognition (CVPR)},
  year={2023}
}

@InProceedings{hoyer2022hrda,
  title={{HRDA}: Context-Aware High-Resolution Domain-Adaptive Semantic Segmentation},
  author={Hoyer, Lukas and Dai, Dengxin and Van Gool, Luc},
  booktitle={Proceedings of the European Conference on Computer Vision (ECCV)},
  pages={372--391},
  year={2022}
}

@article{menze2014multimodal,
  title={The multimodal brain tumor image segmentation benchmark (BRATS)},
  author={Menze, Bjoern H and Jakab, Andras and Bauer, Stefan and Kalpathy-Cramer, Jayashree and Farahani, Keyvan and Kirby, Justin and Burren, Yuliya and Porz, Nicole and Slotboom, Johannes and Wiest, Roland and others},
  journal={IEEE transactions on medical imaging},
  volume={34},
  number={10},
  pages={1993--2024},
  year={2014},
  publisher={IEEE}
}

@article{yao2022darunet,
  title={A novel 3D unsupervised domain adaptation framework for cross-modality medical image segmentation},
  author={Yao, Kai and Su, Zixian and Huang, Kaizhu and Yang, Xi and Sun, Jie and Hussain, Amir and Coenen, Frans},
  journal={IEEE Journal of Biomedical and Health Informatics},
  year={2022},
  publisher={IEEE}
}

@InProceedings{hoyer2022daformer,
  title={{DAFormer}: Improving Network Architectures and Training Strategies for Domain-Adaptive Semantic Segmentation},
  author={Hoyer, Lukas and Dai, Dengxin and Van Gool, Luc},
  booktitle={Proceedings of the IEEE/CVF Conference on Computer Vision and Pattern Recognition (CVPR)},
  pages={9924--9935},
  year={2022}
}

@inproceedings{myronenko20183d,
  title={3D MRI brain tumor segmentation using autoencoder regularization},
  author={Myronenko, Andriy},
  booktitle={International MICCAI brainlesion workshop},
  pages={311--320},
  year={2018},
  organization={Springer}
}

@article{chen2023sam,
  title={Sam fails to segment anything?--sam-adapter: Adapting sam in underperformed scenes: Camouflage, shadow, and more},
  author={Chen, Tianrun and Zhu, Lanyun and Ding, Chaotao and Cao, Runlong and Zhang, Shangzhan and Wang, Yan and Li, Zejian and Sun, Lingyun and Mao, Papa and Zang, Ying},
  journal={arXiv preprint arXiv:2304.09148},
  volume={2},
  number={5},
  pages={7},
  year={2023}
}

@inproceedings{vu2019advent,
  title={Advent: Adversarial entropy minimization for domain adaptation in semantic segmentation},
  author={Vu, Tuan-Hung and Jain, Himalaya and Bucher, Maxime and Cord, Matthieu and P{\'e}rez, Patrick},
  booktitle={Proceedings of the IEEE/CVF conference on computer vision and pattern recognition},
  pages={2517--2526},
  year={2019}
}

@article{bui2023semi,
  title={Semi-supervised learning with fact-forcing for medical image segmentation},
  author={Bui, Phuoc-Nguyen and Le, Duc-Tai and Bum, Junghyun and Kim, Seongho and Song, Su Jeong and Choo, Hyunseung},
  journal={IEEE Access},
  volume={11},
  pages={99413--99425},
  year={2023},
  publisher={IEEE}
}

@inproceedings{milletari2016v,
  title={V-net: Fully convolutional neural networks for volumetric medical image segmentation},
  author={Milletari, Fausto and Navab, Nassir and Ahmadi, Seyed-Ahmad},
  booktitle={2016 fourth international conference on 3D vision (3DV)},
  pages={565--571},
  year={2016},
  organization={Ieee}
}

\end{document}